# COMPARATIVE ANALYSIS OF ADVANCED FEATURE MATCHING ALGORITHMS IN CHALLENGING HIGH SPATIAL RESOLUTION OPTICAL SATELLITE STEREO SCENARIOS

*Qiyan Luo\*, Jidan Zhang\*, Yuzhen Xie, Xu Huang†, Ting Han*

School of Geospatial Engineering and Science, Sun Yat-sen University, Zhuhai, 519082, China

**ABSTRACT**

Feature matching determines the orientation accuracy for the High Spatial Resolution (HSR) optical satellite stereos, subsequently impacting several significant applications such as 3D reconstruction and change detection. However, the matching of off-track HSR optical satellite stereos often encounters challenging conditions including wide-baseline observation, significant radiometric differences, multi-temporal changes, varying spatial resolutions, inconsistent spectral resolution, and diverse sensors. In this study, we evaluate various advanced feature matching algorithms for HSR optical satellite stereos. Utilizing a specially constructed dataset from five satellites across six challenging scenarios, HSROSS Dataset, we conduct a comparative analysis of four algorithms: the traditional SIFT, and deep-learning based methods including SuperPoint + SuperGlue, SuperPoint + LightGlue, and LoFTR. Our findings highlight overall superior performance of SuperPoint + LightGlue in balancing robustness, accuracy, distribution, and efficiency, showcasing its potential in complex HSR optical satellite scenarios.

## 1. INTRODUCTION

Feature matching is a fundamental problem in computer vision applications, which aims to extract feature with significant structures in image pairs and establish feature-to-feature matches [1]. In the field of remote sensing, feature matching is a key step in important applications such as 3D reconstruction [2, 3], change detection [4, 5], and positioning and navigation [6, 7]. The stability and accuracy of feature matching significantly impacts the effectiveness of these downstream applications, especially in image orientation tasks [8].

High spatial resolution (HSR) optical satellite imagery, characterized by its global, large-scale, repetitive observations, reflects detailed three-dimensional spatial structures and distribution of terrestrial features. This has become a research focus in the field of remote sensing. However, the joint processing of HSR satellite images remains challenging. The technical bottleneck lies in achieving high-accuracy registration of different images, closely related to feature matching. With the development of satellite remote sensing, a massive amount of high-resolution optical satellite images from different temporals, viewing angles, radiometric conditions, and sensors have been accumulated in the same region [9]. These pose difficulties for high-accuracy feature matching due to: 1) Changes in land cover in multi-temporal images, particularly seasonal climatic changes (e.g., snow) leading to significant texture variations; 2) Geometric deformations of targets caused by wide-baseline stereo observations [10]; 3) Large radiometric differences between image pairs due to varying solar elevation angles and shooting times; Scale effects, including inconsistencies in 4) spatial and 5) spectral resolutions; 6) Diverse sensors have different imaging parameters.

Current feature matching methods are mainly divided into two categories: 1) Traditional feature matching operators based on shallow features such as gradients and gray-level encoding, and their improved algorithms (e.g., SIFT [11] and its improved versions [12, 13], KAZE [14]); 2) Deep learning feature matching algorithms based on deep level features and effective matching strategy (e.g., SuperPoint + SuperGlue [15, 16], SuperPoint + LightGlue [15, 17], LoFTR [18]).

Our previous research has shown that deep learning demonstrates greater potential in feature matching under challenging conditions compared to traditional algorithms [19], but the data and evaluation methods are still not comprehensive. Specifically, existing feature matching datasets often focus on indoor and outdoor scenes [20], which are ground-level data. Few datasets for satellite overhead perspectives mainly concentrate on multi-modal image analysis [21, 22]. The construction of these datasets typically involves simple geometric operations as well as changes in illumination, which do not comprehensively reflect the challenging conditions in real-world high-resolution optical satellite imagery. Therefore, to advance the development and application of feature matching algorithms in the field of satellite remote sensing, particularly in satellite photogrammetry, we have collected samples from real

---

\* Equal contribution.
† Corresponding author.

Table 1. Details and samples of HSROSS Dataset.

| Satellite (s) | Size of Pairs | Resolution (Spectral, Spatial) | MS | R | S | WB | SPA | SPE |
|---|---|---|---|---|---|---|---|---|
| WV3 | 10 | RGB, 0.31 m | × | × | × | × | × | × |
| | 10 | RGB, 0.31 m | × | √ | × | × | × | × |
| | 10 | RGB, 0.31 m | × | × | × | √ | × | × |
| | 10 | RGB, 0.31 m | × | × | √ | × | × | × |
| GF7 | 10 | Pan, 0.8 m | × | × | × | √ | × | × |
| | 10 | Pan, 0.8 m Multi-Spec, 2.6 m | × | × | × | × | √ | √ |
| GF2 | 10 | Pan, 0.8 m | × | × | √ | × | × | × |
| WV1 WV2 | 10 | WV1: Pan, 0.50 m WV2: Pan, 0.46 m | √ | × | × | √ | × | × |

Pan: Panchromatic. Multi-Spec: Multi-Spectral.

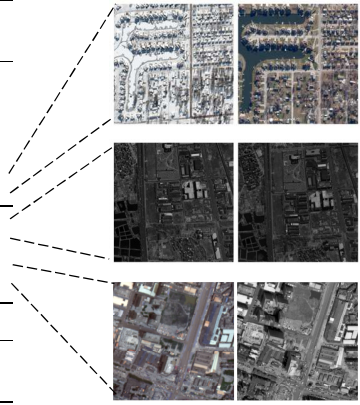

satellite images. We have compiled a dataset from five high spatial resolution optical satellites, covering six challenging conditions, named the High Spatial Resolution Optical Satellite Stereos Dataset (HSROSS Dataset). This dataset has been used to test both the classic computer vision algorithm SIFT and state-of-the-art deep learning feature matching algorithms, serving as a benchmark for our dataset.

The contributions of this study are as follows: 1) We construct HSROSS Dataset, a high-resolution optical satellite image pair dataset for testing feature matching algorithms under various challenging conditions. 2) We comprehensively evaluate performance of the classic method and state-of-the-art deep learning feature matching algorithms on this dataset, from the perspective of feature matching performance and impact on downstream satellite 3D reconstruction tasks.

## 2. DATASET

Notably, satellite data differs significantly from indoor and outdoor datasets [23] in imaging conditions. This distinction is primarily manifested in several aspects of satellite data: 1) Long imaging paths affected by atmospheric conditions; 2) High imaging altitudes causing variations in three-dimensional object shapes; 3) Repetitive observations leading to significant land cover changes between image pairs; and 4) Varying imaging parameters among sensors, affecting spatial and spectral resolutions. However, current satellite datasets provide a limited number of image pairs (e.g., 6 pairs [24]), or they involve data augmentation techniques such as rotation and scaling on a small amount of data [21]. These do not fundamentally increase the diversity of challenging conditions. Moreover, open-source data often lacks corresponding geographic configurations (i.e., rational polynomial coefficient (RPC) files), hampering evaluation in downstream applications like relative orientation and 3D reconstruction.

Therefore, We construct HSROSS Dataset, a high-resolution satellite image dataset under challenging conditions, which includes: 1) Satellite images sourced from five optical high-resolution satellites, including WorldView-1, WorldView-2, WorldView-3, Gaofen-7, and Gaofen-2; 2) Six challenging conditions, i.e., Multi-Source Satellites (MS), Different Radiometric Conditions (R), Seasonal and Land Cover Changes (S), Wide Baseline (WB), Different Spatial Resolutions (SPA), and Different Spectral Resolutions (SPE); 3) Corresponding RPC files and manually selected matching point pairs. Table 1 summarizes the details of each dataset. It also shows some sample data of the dataset. For convenience, in this paper, we use WV to represent WorldView, and GF to represent Gaofen.

## 3. EXPERIMENTS

### 3.1 Implementation details

Our experiments employ pre-trained SuperPoint network [15] provided by Magic Leap, and SuperGlue [16], LightGlue [17] and LoFTR [18] algorithms (LoFTR-DS) are all pre-trained on outdoor dataset MegaDepth [20]. We also use the implementation of SIFT [11] algorithm on GPU. For each pair of data, the central region of 1440 × 1440 size is used during testing. To ensure the fairness of the subsequent comparison, our experiments are inferenced on a single Nvidia GeForce RTX 3090 GPU.

### 3.2 Introduction of algorithms

In the following part, we will simply introduce the algorithms used in the experiments.

**SIFT [11].** It is a notable handcrafted feature descriptor. It detects key points by identifying scale-space extremums. And the feature matching employs a classic method, comparing descriptors from two sets of key points and selecting the optimal matches using distance ratios.

**SuperPoint (SP) [15].** It is a key point extraction algorithm, which realizes self-supervised point of interest detection and descriptor generation through unsupervised training of two pipelines.

**SuperGlue (SG) [16].** This algorithm specializes in feature matching. It assesses and aligns each pair of key points by constructing an association matrix based on descriptors and applying the Sinkhorn algorithm to determine the optimal correspondence between points.

**LightGlue (LG) [17].** It brings advancements in location coding, match prediction, and training supervision. Its adaptability in depth and width stems from two features: the ability to stop inference early when matches are predicted, and excluding unmatched points from later processing.

**LoFTR (LO) [18].** It is a local feature matching algorithm with two phases: coarse dense matching in a low-dimensional space, followed by fine matching with a focus on high-confidence pairs and detailed features in a higher-dimensional space. It also incorporates a distinctive dual-attention mechanism.

### 3.3 Evaluation metrics

Here, we design the following 6 metrics to evaluate the performance of algorithms on HSROSS Dataset.

**Feature Retrieval Ratio (FRR).** This metric assesses the suitability of algorithm-extracted matching points for relative orientation. Ensuring result quality, we considered it is invalid case when there are less than 5 effective matching points or relative orientation accuracy exceeding 10 pixels.

**Number of Correct Matches (NCM).** It quantifies the exact number of accurately matched points, indicating the quality of point extraction useful for applications like relative orientation or regional network adjustment (Eq. 1).

$$CP(x): dist(x_i, EL) \leq \varepsilon \quad (1)$$

Where, $dist(x_i, EL)$ denotes the distance from point $x_i$ to the epipolar line ($EL$), and $\varepsilon$ ($\varepsilon = 3$ in this study) is a threshold.

**Matching Precision (MP).** It calculates the ratio of correct matches to the total matches, presented as a percentage (Eq. 2).

$$MP = \frac{NCM}{PN} \times 100\% \quad (2)$$

Where, $NCM$ is the count of accurate matches, and $PN$ is the total number of matches.

**Normalized Inter-Block Variance (NIBV).** It measures the uniformity of accurate matching point distribution across regions. It involves dividing the image into $N \times N$ blocks ($N = 3$ in this study), calculating the normalized count of accurate points in each block, and then determining the normalized inter-block variance (Eq. 3 & 4).

$$R_i = \frac{NCM_i}{NCM} \quad (3)$$

$$NIBV = \frac{1}{N \times N} \sum_{i=1}^{N \times N} \left(R_i - \frac{1}{N \times N}\right)^2 \quad (4)$$

Where, in Eq. 3, $NCM_i$ is the count of accurate matches in the block $i$, and $NCM$ is the total count of accurate matches in the image. In Eq.4, $N \times N$ is the total number of blocks.

**Accuracy of Relative Orientation (ARO).** It assesses how effectively the matches of algorithm facilitate downstream tasks like 3D reconstruction. Here, a smaller $ARO$ value signifies greater accuracy in the relative orientation model. It's calculated as the average pixel distance between reprojected points and manually selected

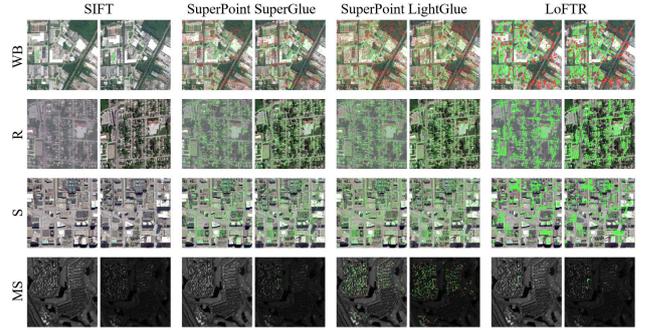

**Fig. 1.** The distribution of matching points obtained by different algorithms in sample data. The red points represent error matches. The green points represent correct matches.

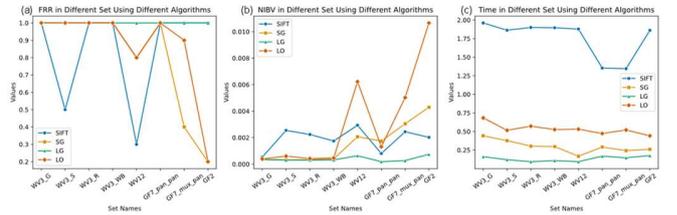

**Fig. 2.** The FRR, NIBV, and time of the different methods on our dataset. (a) FRR; (b) NIBV; (c) Time.

high-accuracy points (Eq. 5).

$$ARO = \frac{1}{n} \sum_{i=1}^{n} dist(x_p, x_i) \quad (5)$$

Where, $n$ is the number of manually selected points, $x_p$ is the reprojection point, and $x_i$ is the manually selected point. The reprojected points are derived through forward-backward projections. Reprojection points are obtained by calculating relative orientation parameters with algorithm-generated matching points, producing object square points of manually selected points through forward intersection, and back-projecting them.

**Time.** This indicator represents the total time for the algorithm to extract all matching points from the 1440×1440 image and complete the feature matching.

## 4. RESULTS AND ANALYSIS

In this study, for four different feature extraction algorithms, we perform a comprehensive evaluation of their performance on our dataset (Table 2).

### 4.1 FRR

The evaluation of feature point extraction algorithms reveals the LG's stability, consistently achieving a 100% success rate across all datasets (Figure 2(a)). While the SIFT, SG, and LO algorithms encounter extraction difficulties under complex conditions such as S, MS, WB, SPA, and SPE.

### 4.2 NCM and MP

The NCM and MP are two critical metrics for evaluating the accuracy of an algorithm's key point extraction process.

Table 2. The NCM, MP, and ARO of the different methods in our dataset. The bold numbers are the best results in that set.

| | Satellite (s) | WV3 | | | | WV1 / 2 | GF7 | | GF2 |
|---|---|---|---|---|---|---|---|---|---|
| | Included Condition (s) | - | S | R | WB | MS, WB | WB | SPA, SPE | S |
| NCM | SIFT | 771.60 | 43.200 | 91.900 | 32.000 | 10.000 | 225.70 | 40.800 | 37.600 |
| | SG | 2697.7 | 1943.5 | 1960.0 | 1211.0 | 26.125 | 45.100 | 14.400 | 3.2000 |
| | LG | 3448.7 | **2692.4** | 2912.5 | 2116.4 | **267.50** | **876.80** | **522.90** | **144.10** |
| | LO | **9243.2** | 1964.3 | **5054.1** | **2547.6** | 19.286 | 244.50 | 70.000 | 11.300 |
| MP | SIFT | 0.9544 | 0.9679 | 0.9595 | 0.8220 | 0.9630 | 0.9839 | **0.8900** | 0.9321 |
| | SG | 0.9700 | 0.9229 | 0.9695 | 0.9206 | 0.8165 | 0.8841 | 0.7717 | 0.3611 |
| | LG | 0.8904 | 0.8120 | 0.8727 | 0.7689 | 0.6399 | 0.7754 | 0.5913 | 0.6211 |
| | LO | **0.9943** | **0.9822** | **0.9928** | **0.9849** | **1.0000** | **0.9876** | 0.8128 | **1.0000** |
| ARO | SIFT | 0.2475 | 0.5471 | 0.7259 | 0.7205 | 1.2187 | **0.3098** | 0.6702 | **0.4312** |
| | SG | 0.2156 | 0.5280 | **0.2843** | 0.3921 | **0.6833** | 0.7831 | 1.1491 | 1.6146 |
| | LG | 0.3023 | 0.4861 | 0.3365 | 0.7738 | 0.7417 | 0.4009 | 0.7518 | 0.6406 |
| | LO | **0.1988** | **0.4594** | 0.2898 | **0.2855** | 2.1069 | 0.6396 | 0.7801 | 0.6102 |

Both metrics collectively provide a comprehensive assessment of the matching accuracy in the algorithm. In Figure 1 and Table 2, the SIFT shows limiting potential in accurate matching, especially under MS and WB conditions, averaging only 10 matches. But it has a notable 90% precision rate of point extraction, due to its scale and rotation invariance. What's more, under the condition of R and MS, the SIFT was unsuccessful as well, because of its inadequate invariance under nonlinear radiation (illumination) conditions. LO algorithm excelled in both match count and precision in RGB imagery, yet its performance obviously declined in panchromatic images. Conversely, SG and LG algorithms achieved more matches but with relatively lower precision, indicating a quantity-quality trade-off.

### 4.3 NIBV

The analysis of NIBV significantly reflects the uniformity of NCM distribution in the image. A smaller NIBV indicates a more uniform distribution of NCM, which contributes to a more robust adjustment process in the subsequent relative orientation step. Combining with Figure 1 and Figure 2(b), we notice that: 1) The SIFT algorithm exhibits the highest variance in color images, likely due to its fewer match counts amplifying the impact of minor changes; 2) In panchromatic images, the LO algorithm shows the highest variance in feature point distribution, possibly due to its reliance on dense matching and a limited number of points. In addition, LG consistently has the lowest NIBV across all conditions, indicating the most uniform feature point distribution.

### 4.4 ARO

In most conditions, the LO algorithm achieves the highest ARO, followed by LG. Notably, LG demonstrates minimal variation in ARO across different conditions, suggesting that its feature matching approach yields more robust matches in terms of relative orientation, maintaining good performance even under complex imaging conditions.

### 4.5 Time

The LG algorithm exhibits excellent execution efficiency, requiring approximately 0.15 seconds to extract a large number of match points from an image of 1440×1440. This efficiency is attributed to its adaptive design, enhancing memory and computational efficiency, simplifying the training process while improving accuracy. In contrast, the execution times for SG, LO, and SIFT algorithms are approximately two, four, and ten times longer than that of LG for processing the same images (Figure 2 (c)).

## 5. CONCLUSIONS

In this study, we rigorously evaluate feature matching algorithms using the HSROSS Dataset from five satellites under six challenging scenarios. Overall, our analysis reveals: SIFT, while with scale and rotation invariant, struggles with wide baselines, seasonal changes, and varying sensors. LoFTR excels in orientation, particularly with RGB imagery, yet underperforms in panchromatic images due to uneven distribution. Compare to realtively low efficiency of SuperPoint + SuperGlue, SuperPoint + LightGlue demonstrates its potential in complex HSR optical satellite scenarios by effectively balancing robustness, accuracy, point distribution, and efficiency.

## 6. ACKNOWLEDGMENTS

The authors thank Qidan Zhang, and Liangchen Zhu for constructing the dataset. The authors also thank the intelligence advanced research projects activity and Digital Globe for providing the WorldView-3 satellite imagery in the CORE3D public dataset. This work was supported by Guangdong Provincial Special Fund for Science and Technology Innovation Strategy.